\documentclass[12pt]{scrartcl}
\usepackage[T1]{fontenc}							
\usepackage[utf8]{inputenc}
\usepackage{lmodern}

\usepackage{geometry}
\KOMAoption{headings}{small}
\KOMAoptions{parskip=half}
\usepackage{setspace}
	\spacing{1.1}
\usepackage[ddmmyyyy]{datetime}
	
\usepackage{scrpage2}
	\ihead{CITlab}
	\ohead{HTRtS-2015}
	\automark[subsection]{subsection}
\usepackage{enumerate}
\usepackage{url}
\usepackage{array, multirow, dcolumn, booktabs}
\usepackage{rotating}

\usepackage{xcolor, graphicx}

\usepackage{amsfonts, amssymb, amsmath, amsthm}

\newcommand{\citlab}{\textsc{CITlab}}

\newcommand{\nac}{\emph{NaC}}

\DeclareMathOperator{\WER}{WER}
\DeclareMathOperator{\CER}{CER}
\DeclareMathOperator*{\argmin}{arg min}
\DeclareMathOperator*{\argmax}{arg max}

\providecommand{\keywords}[1]{\textbf{\textit{Keywords --- }} #1}

\theoremstyle{definition}
\newtheorem{theo}{Theorem}

\newtheorem{rem}[theo]{Remark}

\usepackage{hyperref}

\begin{document}

\title{\citlab\ ARGUS for historical handwritten documents}
\subtitle{Description of \citlab's System for the HTRtS 2015 Task
: Handwritten Text Recognition on the \textsl{tranScriptorium} Dataset
}
\author{
	Gundram Leifert \and Tobias Strauß \and Tobias Grüning \and Roger Labahn
	\thanks{corresponding author; \citlab, Institute of Mathematics, University of Rostock, Germany
		\newline\{gundram.leifert, tobias.strauss, tobias.gruening, roger.labahn\}@uni-rostock.de
	}
}
\date{April 15, 2015}
\maketitle

We describe CITlab's recognition system for the HTRtS competition attached to the 13. International Conference on Document Analysis and Recognition, ICDAR 2015. The task comprises the recognition of historical handwritten documents. The core algorithms of our system are based on multi-dimensional recurrent neural networks (MDRNN) and connectionist temporal classification (CTC). The software modules behind that as well as the basic utility technologies are essentially powered by PLANET's ARGUS framework for intelligent text recognition and image processing.

\begin{abstract} 
We describe \citlab's recognition system for the HTRtS competition attached to the 13.\,International Conference on Document Analysis and Recognition, ICDAR 2015. The task comprises the recognition of historical handwritten documents.
The core algorithms of our system are based on multi-dimensional recurrent neural networks (MDRNN) and connectionist temporal classification (CTC).
The software modules behind that as well as the basic utility technologies are essentially powered by PLANET’s ARGUS framework for intelligent text recognition and image processing.
\end{abstract}
\keywords{MDRNN, LSTM, CTC, handwriting recognition, neural network}

\section{Introduction}
The International Conference on Document Analysis and Recognition, ICDAR 2015\footnote{\url{http://2015.icdar.org}}, hosts a variety of competitions in that area. Among others, the Handwritten Text Recognition on the \textsl{tranScriptorium} Dataset (HTRtS) competition attracted our attention because we expected CITlab's handwriting recognition software to be able to successfully deal with the respective task.

HTRtS\footnote{\url{http://transcriptorium.eu/~htrcontest}} comprises a task of word recognition for segmented historical documents, see  \cite{htrts2014} for all further details. 
These data consist of page images taken from the Bentham collection, a well-known \textsl{transScriptorium} project dataset.

Our neural networks have basically been used previously in the international handwriting competition OpenHaRT 2013 attached to the ICDAR 2013 conference, see \cite{citlab2013openhart}.
Moreover, with a system very similar to the one presented here, the CITlab team also took part in ICFHR's ANWRESH-2014 competition on historical data tables, see \cite{citlab2014anwresh} for the according system description.

Affiliated with the Institute of Mathematics at the University of Rostock, CITlab\footnote{\url{http://www.citlab.uni-rostock.de}} hosts joint projects of the Mathematical Optimization Group and PLANET intelligent systems GmbH\footnote{\url{http://www.planet.de}}, a small/medium enterprise focusing on computational intelligence technology and applications.
The work presented here is part of a common text recognition project 2014 -- 2016 and is extensively based upon PLANET's ARGUS software modules and the respective framework for development, testing and training.

\section{Short Description}
\begin{rem}
This short description is intended for the HTRtS-2015 organizers' information. Here we also explain the abbreviations used in the web form when submitting \citlab\ ARGUS's recognition result files.

\textbf{Please cite this now as:\\
\textsl{private communication, extended version to be published after ICDAR 2015}}.

This draft is preliminary in the sense that it will be further extended to a full paper version. It will be published after the ICDAR 2015 conference when the official final evaluation results are public.
\end{rem}

\subsection{Overview}
Altogether, \citlab\ submits the recognition / transcription results generated by 14 moderately different systems.
While they all mainly rely on our traditional, recurrent neural network based recognition engine ARGUS, the 14 variations arise from combining 2 training schemes, \texttt{trn-1} / \texttt{trn-2}, with 7 decoding schemes, \texttt{dec-BP} / \texttt{dec-CE} / \texttt{dec-DM} and \texttt{dec-E[2|3|4|5]}.
Note that these scheme orderings, suggested by the lexicographic ordering of the respective labels, also reflect both increasing complexity of the schemes, and expected improved quality for the handwritten text recognition task.

\subsection{Basic Scheme}
For the general approach, we may briefly refer to previous \citlab\ system descriptions \cite{citlab2013openhart, citlab2014anwresh, citlab2014htrts} because the overall scheme has essentially not been changed.

\subsection{Preprocessing}
We worked on line polygon images, see \ref{ss:TraDat} for further explanation of the data. Firstly it were applied certain standard preprocessing routines, i.e.
\begin{itemize}\addtolength{\itemsep}{-1.5ex}
	\item image normalization: contrast, size;
	\item writing normalization: line bends, line slope, script slant.
\end{itemize}
Then, images were further unified by \citlab's proprietary writing normalization, thus ensuring a fixed 96px image height with the writing's main body part appropriately placed into and stretched to cover the essential central part of the line image.
These were finally the input images for the subsequent processing with Recurrent Neural Networks (RNN).

\subsubsection{Recurrent Neural Network}
\label{sss:RNN}
The resulting line images were fed into the engine's first core component which we call a Sequence Processing Recurrent Neural Network (SPRNN). Note that we processed entire line images with no further segmentation.

The SPRNN's output then consists of a certain number of vectors. This number is related to the line length because every vector contains information about a particular image position. More precisely, the entries are understood as to estimate the probabilities of every alphabet character at the position under consideration. Hence, the vector lengths all equal the alphabet size, and putting all vectors together leads to the so-called \textsl{confidence matrix}. This is the intrinsic recognition result which will subsequently be used for the decoding.

Note further that, for HTRtS-2015, we worked with the alphabet containing
\begin{itemize}\addtolength{\itemsep}{-1.5ex}
	\item all digits, lowercase and uppercase letters of the standard latin alphabet
	\item special characters /\&£§+-{\textbackslash}\_.,:;!?'"=[]() and \verb*/ /,
		whereby different types of quotation marks and hyphens were mapped to one of the respective symbols.
\end{itemize}
Finally, the above alphabet is augmented by an artificial, non-character symbol, which we denote by \nac. In particular, it may be used to detect character boundaries because, generally speaking, our SPRNNs emit high \nac\ confidences in uncertain situations.

\subsection{Training Schemes}
\citlab\ only participates in the Restricted Track of HTRtS-2015, i.e. for training and testing our systems, we exclusively used data provided within the contest:

\subsubsection{Training Data}
\label{ss:TraDat}
\begin{description}
	\item[\texttt{trn-1}] consists of all \texttt{1stBatch} line polygons, i.e. images of 10\,491 line polygons from 433 pages.
	\item[\texttt{trn-2}] incorporates \texttt{trn-1} and all \texttt{2ndBatch} page images: additional 313 pages, for which the line polygons where not available. Using proprietary \citlab\ tools we extracted 3\,968 more line polygons, s.t.\ altogether, \texttt{trn-2} finally contained 14\,479 training samples.
\end{description}
Note in particular, that from the data provided in HTRtS-2015, we did not use the line images itself because those covered more distortions between adjacent text lines.

\subsubsection{Network Training}
\label{ss:NetTrn}
In both training schemes, various networks have been trained similarly: The number of training epochs slightly varied between 50 and 60, and the decrease of the learning rates was chosen correspondingly.
Moreover, different tries differ in certain hyper-parameters (number of neurons, subsampling rate) and random choices of the initial values for weights that were then optimized by gradient descent procedures.

Out of a larger number of tries, finally 10 networks have been chosen by monitoring the training success on a validation data set which, due to the lack of separate data, was selected from the available training data, see \ref{ss:TraDat}.
Note that the same approach has been used for ranking the 10 final nets in order to choose the best and certain committees, see \ref{ss:DecSch} for details.

\subsection{Decoding Schemes}
\label{ss:DecSch}

\subsubsection{\texttt{Dec-BP}: Best Path decoding}
For decoding the confidence matrix, one starts with the sequence of the most confident character per matrix vector. But in order to get a proper character string over the given alphabet, then two basic transformations have to be applied:
\begin{enumerate}\addtolength{\itemsep}{-1.5ex}
	\item Replace repeated occurrences of the same character by just one!
	\item Delete all \nac\ symbols!
\end{enumerate}
Note first that, due to the order of accomplishing these operations, the special \nac\ symbol serves for distinguishing between proper character repetition vs. just repeatedly seeing the same character while traversing the line image.

Note also that these operations are commonly applied in all decoding schemes! Thus in the following, we know how to proceed from a character sequence from (or \textsl{path} through) the confidence matrix to a valid string interpretation as a required recognition result.

\subsubsection{\texttt{Dec-CE}: \citlab\ Expression decoding}
The details of this decoding developed at \citlab\ will be presented in upcoming publications. Basically it tries to find the most confident string subject to additional restrictions on the internal structure of valid result strings. In HTRtS-2015, the decoded string should be build from expressions which, e.g., look like usual words, have punctuation marks attached to word expressions, have sentences beginning with capital letters \dots\
But note particularly, that this decoding scheme only considers expression syntax -- it does not yet incorporate a dictionary!

\subsubsection{\texttt{Dec-DM}: Dictionary Model decoding}
At this next stage, we include a rather simple language model into the decoding scheme: We try to find the most confident string transcription which belongs to a dictionary. Moreover, besides the string confidences from the recognition result itself, also  word frequencies are taken into consideration. For HTRtS-2015, the dictionary with word frequencies was extracted from the available training data.

\subsubsection{\texttt{Dec-E<n>}: $\mathbf{n}$ Experts Committee decoding}
The above \texttt{Dec-DM} scheme is further extended by simultaneously processing the network output of $n$ different SPRNNs. These were choosen by descending recognition quality on the validation dataset, see \ref{ss:NetTrn}.
For coming to the committee decision, we followed the algorithm proposed in \cite{rover}.
In HTRtS-2015, we submitted four systems with this decoding scheme type, namely for $n\in\{2,3,4,5\}$.

\subsection*{Acknowledgement}

First of all, the \citlab\ team really wishes to express its great gratitude to our long-term technology \& development partner \textit{PLANET intelligent systems GmbH} (Raben Steinfeld, Germany) for the extremely valuable, ongoing support in every aspect of this work. Participating in HTRtS-2015 would not have been possible without that! In particular, we continued using PLANET's software world which was developed and essentially improved in various common \citlab--PLANET projects over previous years.

From PLANET's side, our activities were essentially supported by Jesper Kleinjohann, whom we especially thank for ongoing very helpful discussions and his continuous development support. 

Being part of our current research \& development collaboration project, this work was funded by grant no. KF2622304SS3 (Kooperationsprojekt) in \textsl{Zentrales Innovationsprogramm Mittelstand (ZIM)} by Bundesrepublik Deutschland (BMWi).

Finally, we are indebted to the HTRtS organizers from the PRHLT group at UPV -- in particular Joan Andreu Sánchez -- for setting up this evaluation and the contest as well as the entire \textsl{tranScriptorium} project for providing all the data.

\bibliographystyle{alpha}
\bibliography{SystemDescription}

\begin{thebibliography}{SRTV14}

\bibitem[Fis97]{rover}
J.G. Fiscus.
\newblock A post-processing system to yield reduced word error rates:
  Recognizer output voting error reduction (rover).
\newblock In {\em IEEE Workshop on Automatic Speech Recognition and
  Understanding}, 1997.

\bibitem[LGSL14]{citlab2014anwresh}
Gundram Leifert, Tobias Grüning, Tobias Strauß, and Roger Labahn.
\newblock {CITlab} {ARGUS} for historical data tables: Description of
  {CITlab}'s system for the {ANWRESH-2014} {Word Recognition} task.
\newblock Technical Report 2014/1, Universität Rostock, April 2014.

\bibitem[LLS13]{citlab2013openhart}
Gundram Leifert, Roger Labahn, and Tobias Strauß.
\newblock {CITlab} {ARGUS} for arabic handwriting: Description of {CITlab}'s
  system for the {OpenHaRT} 2013 {Document Image Recognition} task.
\newblock In {\em Proceedings of the NIST 2013 OpenHaRT Workshop [Online]},
  August 2013.
\newblock Available: \url{http://www.nist.gov/itl/iad/mig/hart2013_wrkshp.cfm}.

\bibitem[SGLL14]{citlab2014htrts}
Tobias Strauß, Tobias Grüning, Gundram Leifert, and Roger Labahn.
\newblock {CITlab} {ARGUS} for historical handwritten documents: Description of
  {CITlab}'s system for the {HTRtS} 2014 {Handwritten Text Recognition} task.
\newblock Technical Report 2014/2, Universität Rostock, April 2014.

\bibitem[SRTV14]{htrts2014}
Joan~Andreu Sánchez, Verónica Romero, Alejandro~H. Toselli, and Enrique
  Vidal.
\newblock {ICFHR2014 Competition on Handwritten Text Recognition on
  tranScriptorium Datasets (HTRtS)}.
\newblock In {\em {Proceedings of the International Conference on Frontiers in
  Handwriting Recognition -- ICFHR 2014}}, August 2014.

\end{thebibliography}

\end{document}